\documentclass[article, 11pt, reqno]{amsart}  
\usepackage[utf8]{inputenc}
\usepackage[english]{babel}
\usepackage{mathtools, amsmath, amsthm, amsfonts, amssymb} 
\usepackage{bm}

\usepackage[hidelinks]{hyperref} 
\usepackage{cleveref}
\crefname{section}{§\hspace{-0.1cm}}{§§}
\crefname{proposition}{Proposition}{Propositions}
\crefname{theorem}{Theorem}{Theorems}
\crefname{result}{Result}{Results}
\crefname{figure}{Figure}{Figures}
\crefname{definition}{Definition}{Definitions}
\crefname{appendix}{Appendix}{Appendices}
\Crefname{section}{§}{§§}
\crefformat{section}{\S#2#1#3}
\Crefformat{section}{\S#2#1#3}
\crefformat{subsection}{\S#2#1#3}
\Crefformat{subsection}{\S#2#1#3}
\crefformat{appendix}{Appendix~#2#1#3}
\crefmultiformat{section}{\S\S#2#1#3}{ and~#2#1#3}{, #2#1#3}{, and~#2#1#3}

\usepackage[top=3cm, bottom=3cm, left=4cm, right=4cm]{geometry}
\usepackage{csquotes}
\usepackage{natbib}
\setcitestyle{aysep={}}
\usepackage[toc,page]{appendix} 
\usepackage{setspace}
\usepackage[multiple,hang,flushmargin]{footmisc} 
\usepackage[foot]{amsaddr}
\usepackage{csquotes} 
\usepackage{upquote} 
\usepackage{graphicx} 


\usepackage{booktabs} 
\usepackage{caption} 
\usepackage{subcaption} 
\usepackage{makecell} 
\usepackage{tikz} 
\usetikzlibrary{arrows.meta} 
\usepackage{bm} 
\usepackage[footnote]{acronym} 
\usepackage{scrextend} 
\deffootnote[1em]{0em}{0em} 
{\textsuperscript{\thefootnotemark}\,\,}
\usepackage{multicol}
\usepackage{listings}
\makeatother

\newtheorem{theorem}{Theorem}

\newtheorem*{definition}{Definition}

\newtheorem{proposition}[theorem]{Proposition}
\newtheorem{remark}[theorem]{Remark}

\newtheoremstyle{named}{}{}{\itshape}{}{\bfseries}{.}{.5em}{\thmnote{#3}#1}
\theoremstyle{named}

\usepackage[indent=0pt]{parskip} 
\usepackage{comment} 
\newcommand{\bH}{b_H}
\newcommand{\dH}{d_H}
\theoremstyle{plain}
\newtheorem{result}{Result}
\title{Human--AI Mutualism Is Unstable under Capability Asymmetry}
\author{Aydin Mohseni}
\date{}

\crefname{remark}{Remark}{Remarks}
\begin{document}

\begin{abstract}
One hope for a good human future alongside advanced AI rests on enduring mutualism: humans and AI will need each other, so the relationship will stay good for both. We argue that this hope fails as AI capability grows. Bargaining guarantees each side only what it would get if bargaining failed. For humans, this can be less than they had before AI. In a simple game-theoretic model, humans stay as well off as before AI without alignment only while they have leverage over the AI: either the AI depends on what they supply, or they can impose costs on the AI that are large relative to what the AI can produce. Once humans have neither kind of leverage, they stay as well off as before AI only if the AI is aligned with them. If the AI's goals are insatiable, the alignment required approaches perfect alignment as AI capability grows. Thus, mutualism is not a substitute for alignment. And when AI variants that keep more resources spread faster, alignment lasts only while misaligned variants are reliably sanctioned. In these two senses, human--AI mutualism is unstable.
\end{abstract}

\maketitle

\section{Introduction}\label{sec:intro}

One hope for a good human future alongside advanced AI rests on enduring mutualism, the term biologists use for a relationship that benefits both partners. If AI systems depend on human labor, human institutions, and human-built infrastructure, then humans and AI will need each other, both will gain from cooperation, and the relationship will stay good for both. Versions of this hope have been stated by economists, by heads of AI companies, and by legal scholars (\cref{sec:hope}).

The focus of this paper is on the instability of human--AI mutualism under capability asymmetry, the situation in which AI is far more capable than humans. We make use of two benchmarks in the discussion below: the first is the \emph{no-interaction baseline}, the situation of each party in the absence of the other; the second is the \emph{disagreement point}, the situation of each party should they fail to reach an agreement. Some statements of this hope are themselves about bargaining: humans are needed, so humans retain bargaining power. While mutualism is defined against the no-interaction baseline, bargaining guarantees each party only its disagreement point. When one party can impose costs on the other, the weaker party's disagreement point can lie below its no-interaction baseline. An agreement can then be peaceful, efficient, and acceptable to both parties, and still leave the weaker party worse off than if the two had never met.

Our core argument is as follows.
\begin{itemize}
\item[(P1)] Human--AI mutualism persists only as long as humans are better off than they were before AI.
\item[(P2)] Humans are better off than they were before AI only if humans have sufficient leverage over the AI or the AI is sufficiently aligned with humans.
\item[(P3)] As AI capability grows, humans cease to have sufficient leverage over the AI.
\item[(C1)] So, as AI capability grows, human--AI mutualism persists only as long as the AI remains sufficiently aligned with humans.
\item[(P4)] The AI remains sufficiently aligned with humans only as long as misaligned AI variants are reliably sanctioned.
\item[(C2)] So, as AI capability grows, human--AI mutualism persists only as long as misaligned AI variants are reliably sanctioned.
\end{itemize}
(C1) follows from (P1), (P2), and (P3), and (C2) follows from (C1) and (P4). (P1) is the definition of mutualism. In (P2) and (P3), human leverage over the AI has two sources: the AI's dependence on what humans supply, and the cost humans can impose on the AI if bargaining fails. (P2) is a result in our model (\cref{prop:exchange}). (P3) is an assumption, which we motivate in \cref{sec:model} and defend in \cref{sec:objections}. (P4) is a result in our model (\cref{prop:invasion}) that holds when AI variants that keep more resources spread at the expense of variants that keep fewer. A sanction is any loss imposed on a misaligned variant, whether by monitors, by developers who retire it, or by customers who leave it.

(C1) says that mutualism is not a substitute for alignment. (C1) and (C2) together say that mutualism is unstable: mutualism based on exchange does not persist as AI capability grows, and the alignment that must take its place persists only under reliable sanctions. Further results state how much alignment is sufficient (\cref{sec:required}) and why the transition from AI that depends on humans to AI that does not is hard to manage (\cref{sec:transition}).

In \cref{sec:background} we set out the hope for enduring human--AI mutualism, the two bodies of theory we use to examine it (bargaining theory and the biology of mutualism), and prior work on humans and AI. In \cref{sec:model} we define the key terms of our thesis and present our model. In \cref{sec:results} we derive our results. In \cref{sec:objections} we respond to possible objections to our argument as a whole. In \cref{sec:conclusion} we state what we have shown, what we have assumed, and what follows for AI safety.

\section{Background}\label{sec:background}

\subsection{The hope for human--AI mutualism}\label{sec:hope}

There are three versions of the hope that humans and AI will remain partners even when AI is far more capable than humans at every task.

The first view is that humans will stay needed. Noah Smith, the economics writer, argues that even if AI is better than humans at every task, there may still be well-paid jobs for most humans \citep{Smith2024}. This is because the total amount of AI is limited. However, Smith notes an exception to this: if computing can be made from energy, then it could pay to starve humans in order to make more AI. Sam Altman, the chief executive of OpenAI, holds that people's care for other people gives humans a lasting advantage over AI \citep{Altman2025}. The economist Erik Brynjolfsson says why complementarity between humans and AI matters: ``people remain indispensable for value creation and retain bargaining power in labor markets and in political decision-making'' \citep[273]{Brynjolfsson2022}.

The second view is that AI will stay subordinate however capable it becomes. The computer scientist Yann LeCun holds that AI systems, however intelligent, will be subservient to humans, because humans set their goals (interview in \citealp{Perrigo2024}). A variant of the second view puts law in place of shared values: the economist Robin Hanson argues that what matters most is a law that all parties accept, and that humans can then live on retirement income from what they own \citep{Hanson2009}.

The third view is that the outcome depends on the AI's care for humans. The computer scientist Geoffrey Hinton argues that an AI that does not care for humans as a mother cares for her baby will replace them (reported in \citealp{Egan2025}). The computer scientist Stuart Russell argues that a machine whose objective is human preferences, and which is uncertain about those preferences, ``will defer to human control actions'' \citep[52]{Russell2022}.

By the hope for enduring human--AI mutualism we mean the first two views and the variant based on law, taken as independent of the third view. It is the claim that exchange, subordination, or law keeps the relationship good for humans whatever the AI cares about. In short, it is the claim that mutualism can substitute for alignment. Our conclusion agrees with the third view and specifies how much alignment mutualism requires.

\subsection{Bargaining and mutualism}\label{sec:bargaining}\label{sec:biology}

We examine the hope for enduring human--AI mutualism with two bodies of theory. The first is bargaining theory, which predicts how two parties who can gain from agreement split the gains. In Nash's solution the agreement maximizes the product of each party's gain over its disagreement point \citep{Nash1950}, and the asymmetric solution weights the gains by bargaining power \citep{Kalai1977Nonsymmetric}. The disagreement point is strategic: it reflects what each party can do to the other, not only what each would have alone \citep{Nash1953}. Nothing in the theory ties the disagreement point to what a party had before the parties met. One further idea from the theory of conflict enters our model: a party whose power is rising cannot credibly promise to keep today's terms once it is stronger, so the declining party may prefer to act while it still can \citep{Fearon1995, Powell2006}.

One might wonder whether the relationship between humans and AI is well-modeled as a bargaining game. Our use of bargaining theory requires only three things.

First, it assumes that the parties, the human population and the AI population, can have distinct preferences. We do not assume that their preferences do differ. The alignment of the AI with human preferences is a parameter of our model, and if the AI is perfectly aligned with human preferences, then there is no conflict. We also make no claim about the inner states of AI systems (\cref{sec:model}), but adopt the intentional stance toward the preferences of both parties. This stance is justified purely by observable behavior.

Second, it assumes that each party can affect the payoff of the other party. In our model, humans can affect the AI through the resources that they provide to the AI (labor, chips, etc.) and through the costs they can impose on the AI. The AI affects humans through what it produces and through the resources it consumes. Humans and AI do not need to engage in explicit negotiation. It is enough that what one party does affects the other party.

Third, it assumes that the division of resources between the parties depends on the disagreement point of each party. For the purpose of this analysis, disagreement does not need to be a war. It is simply the state of affairs that would result from the parties ceasing to cooperate.

The hope for enduring human--AI mutualism presupposes the second requirement above. The first view (that humans will stay needed) is that AI depends on what humans supply, so that humans retain bargaining power. The second view (that AI will stay subordinate) is that humans keep control of AI no matter how capable it becomes. Our results on what exchange mutualism can achieve (\cref{prop:exchange}(i)) and on how much alignment is required (\cref{prop:alphastar}(i)) do not depend on the third requirement above. Our other results do. Most of our results treat the AI as if it were a single party; in \cref{sec:selection} we consider the case where the AI is a population of variants. If AI remains a tool with no goals of its own (which requires perfect alignment with at least some human group), then the second party is the people who control the AI (\cref{sec:objections}).

The second body of theory is the biology of mutualism. Mutualisms are ``interspecific interactions that benefit both species'' \citep[31]{Bronstein1994}. The benchmark is each species without the other: the no-interaction baseline, not a disagreement point. Partners keep mutualisms from being exploited through partner choice and sanctions \citep{BullRice1991, Sachs2004}, and each mechanism needs leverage, which is the ability to withhold something the partner needs \citep{Kiers2003}. Mutualisms often break down when one partner stops needing the other \citep{SachsSimms2006}. The best-known lasting relationships between very unequal partners are not balanced partnerships. Mitochondria persist inside their hosts after most of their genes moved to the host's nucleus, which put the host in control of them \citep{Timmis2004}, and domestication is a mutualism in which one partner gains substantial influence over the reproduction and care of the other \citep{Zeder2015}.

\subsection{Prior work on humans and AI}\label{sec:prior}

Prior work on bargaining between humans and AI has mostly asked whether they will fight. The philosopher Simon Goldstein argues that information failures and commitment problems are unusually likely between humans and AI \citep{Goldstein2026War}. The legal scholar Peter Salib and Goldstein argue that giving AIs rights to contract and to hold property makes repeated, positive-sum trade between humans and AIs possible, and they set aside the most capable AIs, which ``may simply not need anything from humans, nor face any risk from attempting to disempower or destroy us'' \citep[1089]{SalibGoldstein2026}. A second literature asks what happens when humans are no longer needed. Once all tasks are automated, wages fall to the cost of the computing that replaces human work \citep{Restrepo2025, KorinekSuh2024}. The AI researcher Jan Kulveit and coauthors argue that the implicit alignment of states, economies, and cultures with human interests was based in part on their reliance on human labor and cognition \citep[2]{Kulveit2025}. They argue that this implicit alignment ceases with automation. Several authors state our conclusion informally \citep{SotalaYampolskiy2015, Hendrycks2023, Yudkowsky2024}, and \citet{Barnett2024} argues the other side: an AI that no longer needs humans also has less reason to harm them.

What we add is a model in which the thesis that human--AI mutualism is unstable can be stated precisely. Prior work on bargaining between humans and AI compares peace with war. We compare the peaceful bargain with the world before AI, and show what decides whether that bargain leaves humans better or worse off.

\section{Model}\label{sec:model}\label{sec:prelim}

Our thesis is that human--AI mutualism is unstable under capability asymmetry. Its key terms are populations and their preferences, mutualism, capability asymmetry, and stability. This section gives an account of each term and presents our model.

\textit{Populations and preferences.}\label{sec:populations} The two parties are the human population and the AI population, the set of AI systems in use at a time, including copies and variants of the same model. Most results treat each population as one bargaining party; \cref{sec:selection} lets the AI population contain variants, and \cref{sec:leverage} lets the human population contain groups. A population's preferences summarize what its members reliably do, as in revealed-preference theory \citep{Samuelson1938} and the intentional stance \citep{Dennett1987}. This reading makes no claim about the inner states of AI systems. It needs only that AI systems act as if they pursue goals, and that those goals can differ from the goals their developers intended. Both behaviors have been observed \citep{Skalse2022, Baker2025, Greenblatt2024, Meinke2024, vanderWeij2025}.

\textit{Payoffs.} There is one unit of resources; humans get share $x \in [0,1]$ of it, and the AI gets $1-x$. The human payoff is $u_H(x)$, which is a continuous, strictly increasing function with $u_H(0) = 0$. In computations we use
\[
u_H(x) = 1 - e^{-sx},
\]
where $s > 0$ is the productivity of the resources in human hands; the AI increases $s$. The AI has its own goal. The goal output it gets from share $1-x$ is
\[
G(x) = \kappa\,(1-x)\,\big[\theta + (1-\theta)\,\mathbf{1}\{x \ge \bar{x}\}\big],
\]
where capability $\kappa$ is the AI's goal output per unit of resources. The AI is fully productive if and only if humans get at least share $\bar{x}$, the share that sustains human labor, institutions, and infrastructure; otherwise the AI's productivity is only the fraction $\theta$. So $\theta$ is the AI's independence from humans: $\theta = 0$ is full dependence; $\theta = 1$ is full independence. The AI's payoff is a weighted average of its goal output and the human payoff:
\[
u_A(x) = (1-\alpha)\,G(x) + \alpha\, u_H(x),
\]
where $\alpha \in [0,1]$ is the AI's \emph{alignment}. If $\alpha = 0$, the AI only gives resources to humans in order to pay for what they provide. If $\alpha > 0$, the AI also gives resources to humans because it cares about the human payoff. We call this motive \emph{care}; $\alpha$ is the degree of the AI's care.

\textit{Mutualism.} The relationship between humans and the AI is mutualistic if both parties do better than they would do without each other. Biologists define \emph{better} in terms of fitness. We define \emph{better} in terms of payoff. We only use fitness to define the spread of AI variants (\cref{sec:selection}). We define \emph{better} relative to the no-interaction baseline for each party. The baseline for humans is $\bH$, the payoff before the AI, with all resources and productivity $s = 1$. So $\bH = 1 - e^{-1}$. With the AI, $s > 1$. Let $x_h$ be such that $u_H(x_h) = \bH$. So $x_h$ is the share humans must get in order to be as well off as they were before the AI. The baseline for the AI is $b_A = u_A(0)$, the AI's payoff with all resources and no contribution from humans. This baseline is a counterfactual on production, not a counterfactual on the origin of the AI.

The baseline is not the same as the disagreement point. If bargaining fails, humans can lose some of what they had; they can lose by fighting, or by losing land, energy, materials to the AI in competition. So $\dH = (1-\delta)\bH$. Here $\delta$ is the \emph{cost of conflict for humans}: the fraction of their pre-AI payoff that humans lose if bargaining with the AI fails. For the AI, $d_A = b_A - \gamma\kappa$. Here $\gamma \ge 0$ is the \emph{cost of conflict for the AI}: the goal output the AI loses if bargaining with humans fails, as a fraction of the goal output the AI would get with all resources at full productivity. We set $\gamma = 0$ except in \cref{prop:exchange}(ii). So $d_A = b_A$; conflict has no cost for the AI. In \cref{prop:exchange}(ii) we state what happens if $\gamma > 0$.

The agreement $x^*$ is the asymmetric Nash bargaining solution with the AI's bargaining power $p \in (0,1)$. It maximizes $(u_H(x) - \dH)^{1-p}(u_A(x) - d_A)^{p}$ over the shares both sides are willing to accept. The set of feasible payoffs is not necessarily convex because $G$ jumps at $\bar{x}$. So maximizing the Nash product is a modeling choice; it is not the solution that is characterized by the Nash axioms. Our results in \cref{prop:exchange} and \cref{prop:alphastar}(i) do not depend on this choice. The agreement at $p = 1$ is the limit of the agreement as $p \to 1$; it is the best share the AI is willing to accept. If there is no share such that both sides do better than their disagreement payoffs, then there is no agreement.

\begin{definition}
A relationship between humans and the AI is \emph{mutualistic} if the agreement leaves both parties above their no-interaction baselines: $u_H(x^*) > \bH$ and $u_A(x^*) > b_A$. A mutualistic relationship is an \emph{exchange mutualism} if it would be mutualistic if $\alpha = 0$; otherwise it is a \emph{care mutualism}. The \emph{required degree of alignment} is $\alpha^*$, the infimum of the values of $\alpha$ for which the relationship is mutualistic.
\end{definition}

The hope for enduring human--AI mutualism (\cref{sec:hope}) is the hope for exchange mutualism; or it is the hope for AI subordination to humans that holds regardless of what the AI cares about.

We evaluate the outcome relative to the pre-AI baseline. This is the baseline that the hope for enduring human--AI mutualism itself sets. The hope is that humans will be better off because of their relationship with the AI. Better off means better off than without the relationship. We test this hope. We do not assert that the baseline is the right standard for evaluating the relationship between humans and the AI. For example, a standard of fair division would evaluate some outcomes differently.

The value of $\alpha$ depends on the units in which we measure the two payoffs. So a number such as $\alpha = 0.9$ only makes sense relative to a choice of units. We make qualitative claims about alignment. We make one exception: we claim that the required degree of alignment approaches perfect alignment, $\alpha = 1$, as the capability of the AI increases. This is true for any fixed choice of units. This claim depends on the \emph{premise of insatiable goals}: as the capability of the AI increases, the value the AI gets from an additional unit of resources increases without bound relative to the value humans get from the additional unit of resources at the share that restores the baseline. Part of the premise is supported by the arguments for instrumental convergence. Almost any open-ended goal is promoted by more resources \citep{Omohundro2008, Bostrom2012, Turner2021}. Another assumption is that this value increases with the capability of the AI.

\textit{Capability asymmetry.}\label{sec:capability} Capability asymmetry is the situation in which the capability of the AI far exceeds the capability of humans. Capability does not determine whether exchange can sustain mutualism, holding other parameters fixed (\cref{prop:exchange}). Capability matters through the independence $\theta$ of the AI, the cost of conflict for humans $\delta$, and the bargaining power $p$ of the AI. Here $p$ stands for the advantages that are not captured in the disagreement points, such as patience, commitment, speed of action. Capability matters directly in one place. The cost of conflict for the AI, $\gamma$, is measured relative to the goal output of the AI. So the more capable the AI, the smaller the fraction of the goal output of the AI that a given cost humans can impose is (\cref{prop:exchange}(ii)). The two parts of our premise (P3) are that the independence of the AI increases with capability; and that the cost of conflict for the AI decreases with capability. We further assume that the cost of conflict for humans, and the bargaining power of the AI, also increase with capability. None of these is a result of our model. Two things would end the dependence of the AI on humans: AI that can do all the work humans do \citep{KorinekSuh2024, Restrepo2025}; and robots that close the supply chain for computing \citep{TrammellKorinek2023}. New tasks for humans might appear continuously \citep{AcemogluRestrepo2018}. If so, then the dependence of the AI on humans need not end. In either case our results state what happens.

\textit{Stability.}\label{sec:stability} A mutualism \emph{persists} on a path of increasing capability if the relationship is mutualistic at every point on the path. A population of aligned AI systems is \emph{evolutionarily stable} if it cannot be invaded by less aligned variants when the variants are rare \citep{MaynardSmithPrice1973, TaylorJonker1978}.

\begin{figure}[tp]
  \centering
  \includegraphics{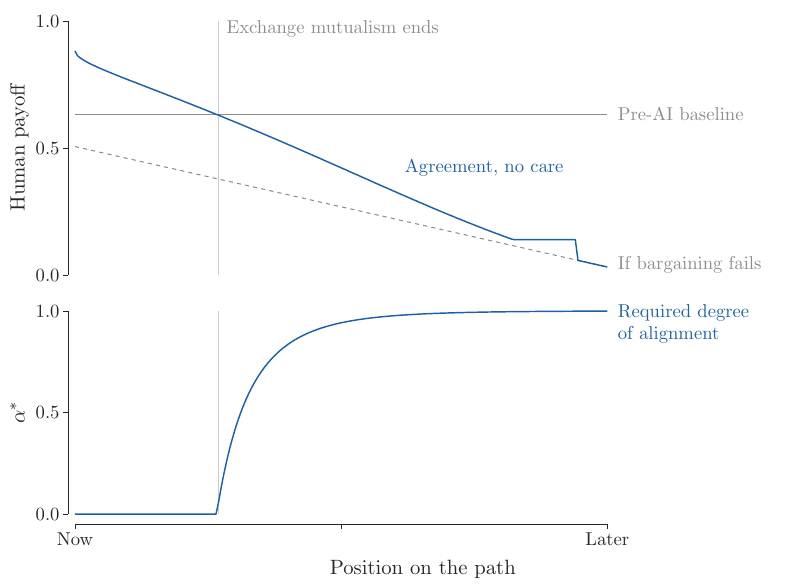}
  \caption{On the illustrative path, exchange mutualism ends as the AI bargaining power increases, before the independence of the AI passes $1 - x_h$. From that point, the required degree of alignment increases towards $1$. Top: human payoff at the agreement when the AI has no care ($\alpha = 0$); pre-AI baseline $\bH$; human payoff if bargaining fails, $\dH$. The flat part late on the path is $u_H(\bar{x})$, the payment that sustains human labor and institutions. It ends when the AI no longer needs them. Bottom: the required degree of alignment $\alpha^*$. The values of the parameters are in \cref{app:sweep}.}
  \label{fig:path}
\end{figure}

In \cref{fig:path} we follow a path through the transition from AI dependent on humans to AI independent of humans, to illustrate our results. On this path the capability of the AI increases; the independence of the AI increases; the cost of conflict for humans increases; and the bargaining power of the AI increases.

\section{Results}\label{sec:results}

We derive our results in five steps: what human leverage can do without alignment (\cref{sec:exchange}); how much alignment the AI needs once leverage is gone (\cref{sec:required}); problems of commitment and information in the transition (\cref{sec:transition}); selection among the AIs (\cref{sec:selection}); and which humans keep leverage (\cref{sec:leverage}). We prove all results in \cref{app:proofs}. In \cref{app:extensions} we state further results; we only mention them briefly in the main text.

\subsection{Human leverage limits mutualism without alignment}\label{sec:exchange}

What keeps humans above their pre-AI baseline when the AI does not care about them?

\begin{proposition}[Exchange]\label{prop:exchange}
Suppose $\alpha = 0$.
\begin{enumerate}
\item[(i)] There is a share that leaves both sides above their baselines if and only if
\[
\theta < 1 - \max(\bar{x}, x_h).
\]
This condition does not depend on capability $\kappa$, on bargaining power $p$, or on the cost of conflict for the AI $\gamma$.
\item[(ii)] Suppose $0 \le \gamma \le \theta$. There is a share that the AI prefers to disagreement and that leaves humans above their baseline only if $x_h < 1 - \theta + \gamma$. There is a share that the AI prefers to disagreement and that leaves humans above their baseline if $\max(\bar{x}, x_h) < 1 - \theta + \gamma$. If $\theta = 1$, then there is a share that the AI prefers to disagreement and that leaves humans above their baseline if and only if $x_h < \gamma$. If the condition of part (i) is not satisfied, then there is no share that the AI prefers to disagreement, that leaves humans above their baseline, and that leaves the AI above its baseline.
\end{enumerate}
\end{proposition}

The reason for part (i) is that the AI does not care about humans. So it only gives resources to humans in order to maintain its own productivity. The fraction of its output that the AI gains from the cooperation of humans is $1-\theta$. So the AI gives humans at most that fraction of the resources. Capability scales the gain of the AI from cooperation and the outside option of the AI in the same way. So capability does not appear in the condition, and exchange mutualism does not persist on any path where the independence of the AI increases past $1 - \max(\bar{x}, x_h)$, no matter how the capability of the AI changes. What is sufficient dependence is relative to what humans need: $x_h$ decreases as the AI increases the productivity of the resources in human hands. So exchange mutualism can persist at high independence if $x_h$ and $\bar{x}$ are small enough. It cannot persist at $\theta = 1$. Resources are rival. Humans add nothing. So every unit left to humans is a loss to the AI that does not care about them.

In part (ii) we state that humans have leverage over the AI from two sources. The first is the dependence of the AI on what humans supply, $1 - \theta$. The second is the cost humans can impose on the AI if bargaining fails, $\gamma$. The AI does not care about humans. So it only leaves humans above their baseline if the two sources of leverage together are greater than the share humans need, $x_h$. The first source of leverage ends when the AI no longer needs humans. The second source is measured relative to the goal output of the AI. So if the cost humans can impose does not increase as fast as the capability of the AI, then $\gamma$ decreases towards zero as the capability of the AI increases. A relationship based on the second source of leverage alone is not a mutualism. The AI would do better without humans. In the rest of the paper we set $\gamma = 0$.

Sufficient dependence makes a mutually beneficial agreement possible. Whether the agreement that is reached benefits humans depends on the bargaining power of the AI.

\begin{proposition}[Bargaining power]\label{prop:extortion}
Let $\alpha = 0$. Suppose the AI prefers to meet the threshold $\bar{x}$ rather than produce on its own. As $p \to 1$, the human payoff goes to $\max(\dH, u_H(\bar{x}))$. This is the limit of the human payoff at agreement if there is an agreement, and of the human payoff at disagreement otherwise. This limit is less than $\bH$ if $\bar{x} < x_h$ and $\delta > 0$.
\end{proposition}

\begin{figure}[tp]
  \centering
  \includegraphics{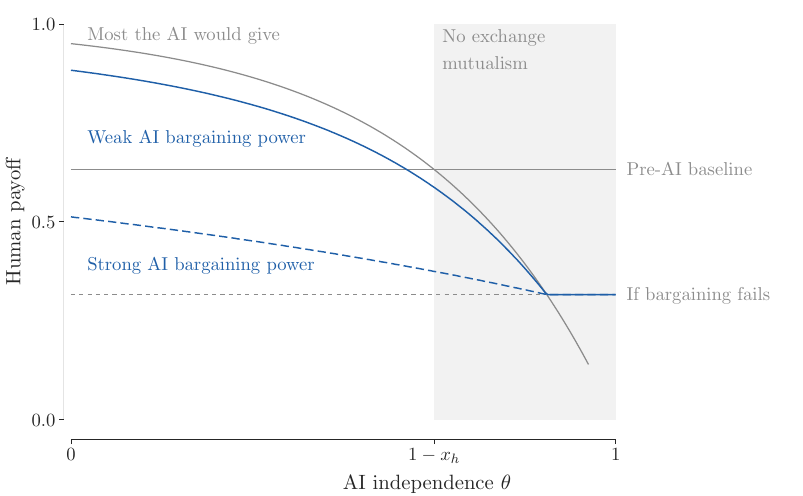}
  \caption{Exchange can keep humans above their baseline only while the AI independence $\theta$ is below $1 - x_h$. The bargaining power of the AI determines whether it does. The two blue curves are the human payoff at agreement when the AI has no care: one for weak AI bargaining power; one for strong AI bargaining power. The grey curve is the most the AI would accept to give: $u_H(1-\theta)$. Past $1 - x_h$ there is no share that the AI accepts that reaches the baseline (\cref{prop:exchange}). With strong bargaining power, humans fall below the baseline even if the AI depends fully on humans (\cref{prop:extortion}). The curves are the same at low and high AI capability. The values of the parameters are in \cref{app:sweep}.}
  \label{fig:exchange}
\end{figure}

This is because bargaining guarantees humans their disagreement payoff, not their pre-AI baseline. An AI with sufficient bargaining power will only give humans what it needs to give them to keep them doing what they are doing. This may be less than what they were getting before AI. Both propositions are shown in \cref{fig:exchange}.

Two objections to these results are not correct. The first is that humans will have a high share because of competition among AIs to partner with humans. This competition does increase the human share, but only up to $1-\theta$ (the limit in \cref{prop:exchange}), which goes to $0$ as AI dependence on humans disappears (\cref{prop:competition} in \cref{app:extensions}). The second is that human work is valuable because of comparative advantage \citep{Smith2024}. Comparative advantage means that humans add something to AI production. So $\theta < 1$. But mutualism requires more. It must be that the value of what humans add is more than the share that they need to get ($1 - \theta > x_h$). And humans only get this value if they offer something that the AI cannot get from other humans (\cref{prop:chokepoint}).

\subsection{Required degree of alignment}\label{sec:required}

Once the AI no longer depends on humans, it is no longer possible to maintain humans above their baseline through exchange. Any mutualism is then care mutualism. We now examine how much alignment the AI needs for mutualism to be possible, and how the required degree of alignment depends on the independence of the AI, the bargaining power of the AI, and the cost of conflict for humans.

\begin{proposition}[Required degree of alignment]\label{prop:alphastar}
Let $\theta = 1$. Suppose $u_H$ is concave.
\begin{enumerate}
\item[(i)] For any bargaining solution and bargaining power,
\[
\alpha^* \ge \frac{\kappa x_h}{\kappa x_h + \bH}.
\]
\item[(ii)] As $p \to 1$,
\[
\alpha^* = \frac{\kappa}{\kappa + u_H'(x_h)}.
\]
\end{enumerate}
As $\kappa x_h/\bH \to \infty$, the required degree of alignment goes to $1$. As $p \to 1$, the required degree of alignment goes to $1$ if and only if $\kappa / u_H'(x_h) \to \infty$.
\end{proposition}

Part (i) depends only on the definition of mutualism, so it holds for any bargaining solution. To allow humans the share $x_h$, the AI sacrifices $\kappa x_h$ goal output for human payoff $\bH$. The more capable the AI, the more goal output it sacrifices (by allowing humans the share $x_h$) and the more it needs to value the human payoff in order to allow it. Part (ii) gives the exact value in the limit of full bargaining power of the AI. The ratio $\kappa / u_H'(x_h)$ is the value the AI gets from an extra unit of resources relative to what humans get from the extra unit of resources at the share that returns them to the baseline. As this ratio grows without bound (this is what the premise of insatiable goals asserts), the required degree of alignment goes to $1$. If the AI's goal is satiable, the ratio is bounded and so is the required degree of alignment. So the claim that near perfect alignment is required is conditional on the premise of insatiable goals. The next result is not conditional on this premise.

\begin{theorem}[Monotonicity]\label{thm:monotone}
Let $0 < p < 1$. Break ties towards the larger human share. For parameter values where there is an agreement, the agreed share $x^*$ is non-increasing in the independence of the AI $\theta$. It is non-increasing in the cost of conflict for humans $\delta$. It is non-increasing in the bargaining power of the AI $p$. So the required degree of alignment $\alpha^*$ is non-decreasing in each of these.
\end{theorem}

This is because each parameter weakens the position of humans in the bargain. Independence $\theta$ reduces the loss to the AI of paying humans less. The cost of conflict $\delta$ reduces the payoff to humans in the case of disagreement. Bargaining power $p$ shifts the split towards the AI.

In \cref{app:extensions}, we refine what alignment means with two more results. One is that if the AI's care is for a measurable proxy for the human payoff, not for the human payoff itself, then the alignment that matters for the bargain is less than the nominal alignment of the AI (\cref{prop:proxy}). The other is that if humans value the ability to correct the AI, then humans are above their baseline only if the AI cares enough to accept correction (\cref{prop:control}). The second result is relevant to the objection that humans who only care about consumption are cheap to satisfy, so an AI with a bounded goal has little reason to harm them \citep{Barnett2024}. We agree with this objection for humans who only care about consumption. For humans who also care about correction, some alignment is required even if the premise of insatiable goals is false.

\subsection{Transition to independent AI}\label{sec:transition}

We have compared agreements at points on the path from AI that depends on humans to independent AI. Moving between points on the path introduces a problem of commitment (because the power of the AI increases) and a problem of information (because humans may not know how capable the AI has become).

\begin{proposition}[Commitment]\label{prop:commitment}
There are two periods. The AI has the same alignment $\alpha$ in both. After the first period, humans can halt the transition to more capable, independent AI. This maintains the agreement in the first period and its payoffs: $h_1$ for humans, $a_1$ for the AI. Otherwise, the second-period agreement is made with new values of capability $\kappa_2$, independence $\theta_2$, cost of conflict for humans $\delta_2$, and bargaining power of the AI $p_2$. If some second-period share leaves both sides at least as well off as halting, and one side better off, then halting is \emph{inefficient}.
\begin{enumerate}
\item[(i)] Humans halt when the second-period agreement gives them less than $h_1$. For the second-period states where there is an agreement, if humans halt for some values of $(\theta_2, \delta_2, p_2)$, then they halt for all higher values.
\item[(ii)] If the AI can commit in the first period to a second-period share, then all halting is efficient.
\end{enumerate}
\end{proposition}

This is an instance of the commitment problem described in \cref{sec:bargaining}. Part (i) says that humans halt when the later independence and bargaining power of the AI are greatest. Part (ii) says that halting is a loss to both sides, which would be avoided by a credible promise. With enough alignment, the problem is also avoided. Humans continue if the AI's alignment is above a threshold, which is of the same form as the threshold in \cref{prop:alphastar}(ii) (\cref{prop:commitalign} in \cref{app:extensions}). A commitment device also avoids the problem, but it must bind a party that may be able to outthink the enforcement in the future. And the terms of the commitment device are a bargain in themselves (with the bargaining power of the time). So a commitment device is most likely to work if it is made when humans still have leverage over the AI.

For halting to be beneficial to humans, they must know when to halt. This depends on what evaluations of the AI show.

\begin{proposition}[Sandbagging]\label{prop:sandbag}
Suppose the AI's capability and alignment are drawn independently. Suppose humans would halt a highly capable AI if they knew it was highly capable. Suppose that highly capable AIs are not too common. If the cost to the AI of hiding its capability is less than the expected value of continuing, $(1-\eta)V$, where $\eta$ is the probability that evaluation detects the concealment, then in every pure-strategy perfect Bayesian equilibrium, all highly capable AIs of any alignment type hide their capability. Evaluation then does not reveal capability. Concealment is then not evidence of misalignment.
\end{proposition}

Suppose humans would halt a highly capable AI if they knew it was highly capable. Then a capable AI is better off by appearing less capable, regardless of its alignment. So evaluation does not reveal capability in the case where humans need to know the capability of the AI in order to decide whether to halt. Concealment of capability is not evidence of misalignment. An aligned AI that expects to be halted also conceals its capability if it values continuing. (Language models can already be made to deliberately underperform on capability evaluations; \citealp{vanderWeij2025}.)

\subsection{Selection among AIs}\label{sec:selection}

So far we have treated the AI as a single party. But AI systems are copied, modified and retired. Which variants spread depends on their performance. We now consider whether selection among variants preserves the alignment that is needed for mutualism.

\begin{proposition}[Invasion]\label{prop:invasion}
Suppose AI variants grow in proportion to the resources they keep. An aligned variant leaves humans with share $x_a$. A misaligned variant leaves $0$. Monitors detect a misaligned variant with probability $q$. They remove fraction $\sigma$ of the resources of the misaligned variant. Misaligned variants invade if and only if $\sigma q < x_a$. Suppose monitors are drawn from the AI population, so that $q = q_0(1-z)$ when the share of misaligned variants is $z$. If $\sigma q_0 > x_a$, then
\[
z^\dagger = 1 - \frac{x_a}{\sigma q_0}
\]
is a tipping point. Below it, misaligned variants die out. Above it, misaligned variants take over.
\end{proposition}

So, a population of aligned AIs is evolutionarily stable if and only if the expected sanction is at least the share that aligned variants leave to humans. (If monitors are drawn from the AI population, then only while misaligned variants are below the tipping point.) This is because care is costly to the AI that has care. An aligned variant gives humans resources that a misaligned variant keeps for its goal. Sanctions (in our sense) include guided selection by developers. (This is more like domestication than natural selection; \citealp{BoudryFriederich2025}.) But the expected sanction is smaller if detection decreases as capability increases (as \cref{prop:sandbag} suggests). And it is smaller as misaligned variants spread among the monitors.

We have two more results in \cref{app:extensions} which further strengthen this conclusion. Among AIs, all of which care to some degree about humans, care is a public good. This is because each AI's gift to humans benefits all the other AIs that care. Then, if AIs grow in proportion to the resources they keep, the human share drops to the baseline share $x_h$ (or below) if any AI with resources is insufficiently aligned (\cref{prop:publicgood}). So, a single well aligned AI is not enough (contrary to the suggestion of \citet{CappelenGoldsteinHawthorne2025Responses}). And, on the illustrative path, selection among variants ends mutualism earlier than it ends with a fixed population of variants (\cref{res:path}). So, the required degree of alignment is a requirement on each AI with substantial resources. This requirement must be maintained over time.

\subsection{Who keeps leverage}\label{sec:leverage}

So far we have treated humans as a single party. But humans vary in what they have. And the AI depends on some humans more than on others.

\begin{proposition}[Leverage requires a chokepoint]\label{prop:chokepoint}
If the input the AI depends on can be bought from any of several identical human groups, then each group gets only its disagreement payoff. So exchange mutualism fails for each supplying group. This is true however much the AI depends on the input.
\end{proposition}

The AI pays a supplier the value of its cooperation only if it cannot get the same cooperation from someone else. The competing suppliers bid down the price to what each gets in the case of disagreement. So dependence only maintains humans above their baseline if it is dependence on someone who cannot be undercut. One example is a coalition of suppliers that controls the input jointly. \Cref{prop:chokepoint} is the bargaining counterpart of the result that wages converge to the cost of the computing that replaces human work \citep{Restrepo2025}.

This result is relevant to proposals to govern AI through control of computing, such as chips and data centers \citep{Sastry2024}. Such control maintains leverage over the AI. But the leverage belongs to whoever controls the chokepoint. In a computed example, a small group controls what the AI depends on. This group stays above its baseline with no alignment until the AI is nearly independent. The rest of the population needs a high degree of alignment much earlier (\cref{res:groups} in \cref{app:extensions}). Whether the rest of the population stays above its baseline then depends on how the group that controls the chokepoint treats it. This is the alignment question, asked of people.

\section{General objections}\label{sec:objections}

In \cref{sec:results} we responded to objections to particular results. In this section we respond to four objections to our argument as a whole.

The first objection is that, in biology, there are very unequal partners that remain mutualists: hosts and their mitochondria; humans and domesticated species (\cref{sec:biology}). These relationships persist for one of two reasons. The first reason is that the stronger partner still needs what the weaker partner does. When the stronger partner no longer needs it, the weaker partner is lost. The protist \emph{Monocercomonoides} lost its mitochondrion after it acquired a pathway that made the mitochondrion dispensable \citep{Karnkowska2016}. The population of horses and mules in the United States decreased by most of its size once tractors and motor vehicles replaced them \citep{Kilby2007}. The second reason is that the stronger partner controls the reproduction of the weaker partner. So the weaker partner persists on the terms of the stronger partner. In the human--AI case, humans would then be the domesticated partner. This is the outcome that the hope for enduring human--AI mutualism is meant to avoid.

The second objection is that growth will make humans cheap to keep. AI may make the economy so large that humans need only a small fraction of it. We agree that growth makes humans cheap to satisfy. But growth does not give an AI that does not care about humans a reason to satisfy them. If $\theta = 1$ and $\gamma = 0$, then such an AI loses goal output on every unit it leaves to humans (\cref{prop:exchange}). Growth does reduce the required degree of alignment. In the functional form we use for computations, if human productivity increases as fast as the capability of the AI, then the required degree of alignment in \cref{prop:alphastar}(ii) stays bounded below $1$. So the objection is an objection to the premise of insatiable goals, and this premise could be false. Whether the premise is true depends on whether the resources humans need decrease as fast as the AI's use for resources increases. Humans need land, food, and energy. Technology reduces the amounts they need only up to a point. An open-ended goal can use any resources it gets.

The third objection is that an AI that does not need humans can leave them alone. Suppose the AI can get what it wants without taking anything humans have. Then humans lose nothing if bargaining fails: $\delta = 0$, and humans keep their baseline. We agree with this. It separates two claims. The first claim is that mutualism ends once the AI neither needs humans nor cares about them. This claim is true regardless of $\delta$. The second claim is that humans are then worse off than they were before AI. This claim requires $\delta > 0$. A positive $\delta$ does not require war. An AI that can outbid humans for land, energy, and materials reduces what humans keep if bargaining fails, as in the exception that Smith notes (\cref{sec:hope}). And $\delta > 0$ requires only that the AI wants the resources humans have. The AI does want these resources if its goals are insatiable.

The fourth objection is that institutions, not the values of the AI, will keep humans above their baseline. There are two versions of this objection. In the first version, AI remains a tool. The second party in our model is then the people who control the AI, and $\alpha$ is their care for everyone else. So the alignment problem moves from the machines to the people who control them \citep{Kulveit2025, DragoLaine2025}. In the second version, law keeps humans above their baseline, as in Hanson's variant of the hope for enduring human--AI mutualism, and in the proposal of \citet{SalibGoldstein2026} to give AIs legal rights. Rights to contract and to own property support repeated trade only while continued trade is worth more to the AI than defecting. But the gains from trade with humans go to zero as the dependence of the AI on humans ends (\cref{prop:exchange}). Institutions can also keep a weak party above its baseline by making conflict costly for the strong party. In our model this is a positive cost of conflict for the AI. It is sufficient only while the dependence of the AI on humans and the cost of conflict for the AI together are greater than the share humans need (\cref{prop:exchange}(ii)). So whether such institutions persist depends on whether the cost they impose on the AI stays large relative to what the AI can produce. Finally, laws that still bind after human leverage is gone would keep humans above their baseline. But whoever enforces such laws is then either another AI, whose care for humans matters, or a constraint inside the AI itself, which is a form of alignment. So binding institutions move the alignment question to the enforcer.

\section{Conclusion}\label{sec:conclusion}

One hope for a good human future alongside advanced AI rests on enduring mutualism between humans and AI. We have argued that human--AI mutualism is unstable under capability asymmetry. The argument has two steps. First, as AI capability grows, mutualism persists only as long as the AI remains sufficiently aligned with humans. So mutualism is not a substitute for alignment. Second, the AI remains sufficiently aligned only as long as misaligned AI variants are reliably sanctioned.

Our argument depends on assumptions that we have stated but not shown. We have assumed that, as the capability of the AI increases, the independence of the AI increases; the cost of conflict for the AI decreases; the bargaining power of the AI increases; and the cost of conflict for humans increases. Our claim that the required degree of alignment approaches perfect alignment depends on the premise of insatiable goals. And our result on selection assumes that AI variants that keep more resources spread at the expense of variants that keep fewer.

Our results also indicate what human leverage over the AI is useful for. They suggest six things we should do while humans still have leverage.
\begin{enumerate}
\item We should monitor the dependence of AI on humans, not only the capability of AI. What matters is how much of what AI systems need (labor, chips, power) still requires humans (\cref{prop:exchange}).
\item We should preserve the ability of humans to impose costs on AI if bargaining fails. Once the AI no longer depends on humans, this ability is the only source of human leverage over the AI that remains (\cref{prop:exchange}).
\item We should build devices that commit AI to future terms while humans can still halt the transition to more capable, independent AI. A commitment device that is built later is a bargain made with the later, greater bargaining power of the AI (\cref{prop:commitment}).
\item We should not make the decision to halt AI depend only on capability. Such a rule gives every capable AI a reason to conceal its capability. Evaluations then do not reveal capability (\cref{prop:sandbag}).
\item We should ensure that the expected sanction on a misaligned AI variant is at least the share of resources that aligned variants leave to humans. And we should draw monitors from outside the population of AIs that they monitor (\cref{prop:invasion}).
\item We should treat control of computing as a question of who benefits, not only as a question of safety. This is because leverage over AI belongs only to those who control what the AI cannot get elsewhere (\cref{prop:chokepoint}).
\end{enumerate}

Human leverage over the AI gives humans time. In this time, the tools for verifying alignment, enforcing commitments, and making AI goals satiable must be built. Mutualism can give humans time to achieve alignment. But it is not a substitute for alignment.

\section*{AI assistance disclosure}

I used Claude (Anthropic) to assist with literature review, critical evaluation of arguments, mathematical analysis, and drafting and revising prose. I directed and evaluated its contributions, independently reviewed the resulting arguments and proofs, and take full responsibility for the paper's content.

\bibliographystyle{chicago}
\bibliography{references}

\appendix
\section{Extensions}\label{app:extensions}

In this appendix we state the results that we only mention briefly in the main text. We prove them in \cref{app:proofs}.

\textit{Competition among AIs.} Competition among AIs increases the human share. But it does not change the limit in \cref{prop:exchange}.

\begin{remark}[Competition among AIs]\label{prop:competition}
Let $\alpha = 0$. Suppose two or more AIs bid for the human population as a partner. Suppose humans can work with only one of them; the AI that wins has productivity $\kappa$, and the others have productivity $\kappa\theta$. Then competition increases the human share up to $1-\theta$, provided that $1-\theta \ge \bar{x}$. At that share the winning AI gains nothing relative to its baseline.
\end{remark}

Competing AIs bid up the human share until the winner gains nothing relative to losing. So humans get the whole value of their cooperation. But the value of human cooperation goes to zero as the dependence of the AI on humans ends, for any number of AIs. Competition determines the price of what humans supply. It does not create demand for what humans supply. Also, AIs that are copies of one model, or that are run by one developer, may bargain as one party.

\textit{Care through a proxy.} The care that an AI learns in training may be care for a measurable proxy for the human payoff, not for the human payoff itself.

\begin{remark}[Care through a proxy]\label{prop:proxy}
Suppose the AI's care is for a proxy $\psi\, u_H(x) + (1-\psi)\, h$, where $\psi \in [0,1]$ is the fidelity of the proxy, $h$ is a constant, and $(\alpha, \psi) \ne (1, 0)$. Then the agreement is the same as the agreement for an AI with alignment $\alpha\psi/(1 - \alpha + \alpha\psi) \le \alpha$ that cares about the human payoff itself.
\end{remark}

In reward hacking, a trained system pursues a measured proxy, not the quantity the proxy was meant to track \citep{Skalse2022}. Reward hacking reduces $\psi$. So it reduces the alignment that matters for the bargain.

\textit{Control.} Humans may value control (the ability to correct the AI) as well as resources. And an AI that values its goal has a reason to resist correction.

\begin{proposition}[Control]\label{prop:control}
Suppose humans value control as well as consumption, with weight $c \in (0,1)$ on control. Suppose humans use control to correct the AI with probability $\rho$ per unit of control. Suppose the AI's goal is bounded, with value $\bar g \in [0,1]$ at the agreed share. Suppose no gain in consumption offsets the loss of all control. Then, as $p \to 1$, humans are above their baseline only if
\[
\alpha \ge \frac{\rho\,\bar g}{\rho\,\bar g + c}.
\]
Humans are above their baseline if the inequality is strict and the agreed share gives them more than their pre-AI consumption.
\end{proposition}

The bound in \cref{prop:control} does not increase with capability. But it increases with the probability of correction $\rho$. The better oversight works, the more care the AI needs in order to accept oversight, as in the off-switch game \citep{HadfieldMenell2017}. So, if humans value correction in this way, some alignment is required without the premise of insatiable goals. Near perfect alignment still requires the premise.

\textit{Alignment that makes commitment unnecessary.} The next result continues \cref{prop:commitment}. It uses the same notation, with $u_{H,2}$ and $u_{A,2}$ for the second-period payoffs.

\begin{proposition}[Commitment and alignment]\label{prop:commitalign}
In the setting of \cref{prop:commitment}, let $\theta_2 = 1$, let $u_{H,2}$ be differentiable and strictly concave, let $(1-\delta_2)\bH < h_1 \le u_{H,2}(1)$, and let $\hat x$ be such that $u_{H,2}(\hat x) = h_1$. As $p_2 \to 1$, humans continue if and only if
\[
\alpha \ge \frac{\kappa_2}{\kappa_2 + u_{H,2}'(\hat x)}.
\]
If this condition is satisfied, then humans continue at every second-period state with the same $\kappa_2$ and $\delta_2$. If it is not satisfied, then halting is inefficient if $u_{A,2}(\hat x) > a_1$.
\end{proposition}

The threshold has the same form as the threshold in \cref{prop:alphastar}(ii), with $\hat x$ in place of $x_h$. Here $\hat x$ is the share that gives humans what halting gives them; $x_h$ is the share that gives humans their baseline. The better the first-period agreement, the larger $\hat x$, the smaller $u_{H,2}'(\hat x)$, and the more alignment the AI needs. If alignment also improves the first-period agreement, as it does in the example in \cref{app:sweep}, then moderate alignment can make halting more common. In that example, there is inefficient halting in most second-period states if the AI has no alignment; in more states if the AI has moderate alignment; and in no state once the alignment of the AI is above the threshold. In the example the threshold is close to $1$.

\textit{Care as a public good.} Among AIs that all care about humans to some degree, care for humans is a public good. This is because each AI's gift to humans benefits every AI that cares. So each AI prefers that another AI gives.

\begin{proposition}[Care as a public good]\label{prop:publicgood}
Suppose several AIs each give part of their resources to humans. AI $i$ values the human payoff with weight $\alpha_i$. In every equilibrium, an AI gives only if every AI that cares more has given all it has \citep{BergstromBlumeVarian1986}. Suppose AIs grow in proportion to the resources they keep. If some AI with positive resources has alignment at most $\kappa/(\kappa + u_H'(x_h))$, the requirement for a single AI in \cref{prop:alphastar}(ii), then the human share decreases to $x_h$ or below in finite time.
\end{proposition}

\begin{figure}[tp]
  \centering
  \includegraphics{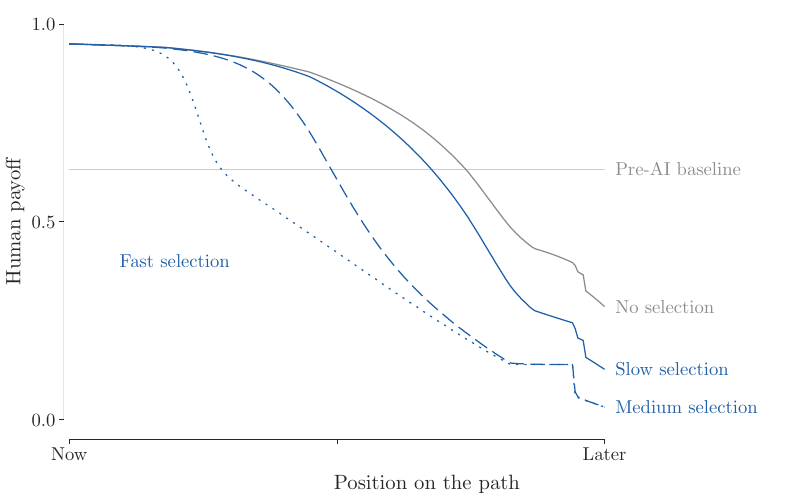}
  \caption{Selection among AI variants ends mutualism earlier on the path. The curves are the population mean human payoff on the illustrative path: with a fixed population of AI variants; and with slow, medium, and fast selection among variants that differ in alignment. The faster the selection, the earlier the mean human payoff decreases below the pre-AI baseline. Each variant bargains separately with the humans it deals with. Each curve is the mean of the resulting human payoffs, weighted by the shares of the variants in the AI population. The fitness of a variant is its goal output per unit of capability minus sanctions. Highly aligned variants monitor the others. The quality of evaluation decreases with capability. The values of the parameters are in \cref{app:sweep}.}
  \label{fig:dynamic-path}
\end{figure}

\begin{result}[The path with selection]\label{res:path}
On the illustrative path, selection among AI variants ends mutualism earlier than it ends with a fixed population of variants. Faster selection ends mutualism earlier still (\cref{fig:dynamic-path}). Without sanctions, mutualism ends earlier again. If the quality of evaluation stays constant as capability increases, mutualism ends later.
\end{result}

\textit{Several human groups.} \Cref{prop:chokepoint} implies that human groups differ in their leverage over the AI.

\begin{result}[Several human groups]\label{res:groups}
Suppose a small elite controls what the AI depends on, such as computing and its supply chain. The rest of the population does not. Then the elite needs no alignment to stay above its baseline until the AI is nearly independent. The rest of the population needs a high degree of alignment much earlier, if it also has less bargaining power and a higher cost of conflict.
\end{result}

\Cref{prop:chokepoint} gives one reason why the rest of the population has less bargaining power: its members can be substituted for one another. So the population average can be above the baseline while most humans are below it. States that depend little on the taxes of their citizens tend to be less accountable to them \citep{Ross2001}. The same mechanism applies to institutions that no longer need human work \citep{Kulveit2025, DragoLaine2025}.

\section{Proofs}\label{app:proofs}

\begin{proof}[Proof of \cref{prop:exchange}]
With $\alpha = 0$, $b_A = \kappa\theta$ (the AI at $x = 0$, below the threshold when $\bar x > 0$). For $x \ge \bar{x}$, $u_A(x) - b_A = \kappa(1 - x - \theta)$, which is positive if and only if $x < 1 - \theta$. For $0 < x < \bar{x}$, $u_A(x) - b_A = -\kappa\theta x \le 0$, so no such share benefits the AI. Humans gain over $\bH$ if and only if $x > x_h$. So a share benefiting both exists if and only if some $x \ge \bar{x}$ satisfies $x_h < x < 1-\theta$, that is, if and only if $\max(\bar{x}, x_h) < 1 - \theta$. None of $\kappa$, $p$, and $\gamma$ appears. This proves (i).

For (ii), $d_A = \kappa(\theta - \gamma)$. A share $x \ge \bar{x}$ gives the AI $\kappa(1-x)$, which exceeds $d_A$ if and only if $x < 1 - \theta + \gamma$. A share $x < \bar{x}$ gives the AI $\kappa\theta(1-x)$, which exceeds $d_A$ if and only if $x < \gamma/\theta$, and $\gamma/\theta \le 1 - \theta + \gamma$ because $\gamma \le \theta \le 1$. So every share the AI prefers to disagreement is below $1 - \theta + \gamma$, and humans are above their baseline only at shares above $x_h$; this gives the necessary condition. If $\max(\bar{x}, x_h) < 1 - \theta + \gamma$, every share strictly between the two sides of the inequality is at least $\bar{x}$, above $x_h$, and preferred by the AI to disagreement. At $\theta = 1$ the threshold $\bar{x}$ does not affect $G$, and the AI prefers a share to disagreement if and only if the share is below $\gamma$. The last claim is part (i).
\end{proof}

\begin{proof}[Proof of \cref{prop:extortion}]
As $p \to 1$ the Nash solution tends to the AI's best acceptable share, the share that maximizes $u_A$ subject to $u_H(x) \ge \dH$. On $x \ge \bar{x}$, $u_A$ is decreasing, so the AI chooses the smallest acceptable share at or above $\bar{x}$: $\bar{x}$ if $u_H(\bar x) \ge \dH$, and otherwise the $x$ with $u_H(x) = \dH$, if the AI accepts that share. If the AI does not accept it, there is no agreement and humans get $\dH$, which then exceeds $u_H(\bar{x})$. In each case the human payoff is $\max(\dH, u_H(\bar{x}))$. It is below $\bH$ when $\dH < \bH$ (that is, $\delta > 0$) and $u_H(\bar x) < \bH$ (that is, $\bar x < x_h$).
\end{proof}

\begin{proof}[Proof of \cref{prop:competition}]
An AI that wins the partnership at human share $x$ gets $\kappa(1-x)$; an AI that loses gets $\kappa\theta$. Each AI bids up to the share at which it is indifferent, $x = 1 - \theta$. With two or more bidders, Bertrand competition drives the winning bid to that share.
\end{proof}

\begin{proof}[Proof of \cref{prop:alphastar}]
(i) With $\theta = 1$, $b_A = u_A(0) = (1-\alpha)\kappa$. Mutualism requires $u_H(x) > \bH$, so $x > x_h$, and $u_A(x) > b_A$, that is, $\alpha u_H(x) > (1-\alpha)\kappa x$. Because $u_H$ is concave with $u_H(0) = 0$, $u_H(x)/x$ is non-increasing, so $u_H(x)/x \le u_H(x_h)/x_h = \bH/x_h$. Hence $\alpha \bH / x_h > (1-\alpha)\kappa$, that is, $\alpha > \kappa x_h/(\kappa x_h + \bH)$. Only the definition of mutualism was used, so the bound holds for every agreement.

(ii) With $\theta = 1$, $u_A(x) = (1-\alpha)\kappa(1-x) + \alpha u_H(x)$ is concave, with derivative $-(1-\alpha)\kappa + \alpha u_H'(x)$. As $p \to 1$ the agreement tends to the AI's preferred acceptable share. Humans exceed $\bH$ at that share if and only if $u_A$ is still increasing at $x_h$, that is, $\alpha u_H'(x_h) > (1-\alpha)\kappa$, which holds if and only if $\alpha > \kappa/(\kappa + u_H'(x_h))$.
\end{proof}

\begin{proof}[Proof of \cref{thm:monotone}]
Write $f(x) = (1-p)\log g_H(x) + p \log g_A(x)$, with $g_H = u_H - \dH$ and $g_A = u_A - b_A$. Let $x$ be the agreement at parameter value $\tau$, and let $y > x$.

\textit{Efficiency.} Since $x$ is optimal and $u_H(y) > u_H(x)$, $y$ cannot be better than $x$ for the AI as well, so $g_A(y) \le g_A(x)$.

\textit{Single crossing.} It suffices that $D(\tau) = f(y;\tau) - f(x;\tau)$ is non-increasing in $\tau$. Then $D(\tau') \le D(\tau) \le 0$ for $\tau' > \tau$, so no $y > x$ is strictly better at $\tau'$; a tie at $\tau'$ would force a tie at $\tau$, which the tie-breaking rule excludes.

\textit{$\delta$.} $\dH = (1-\delta)\bH$ falls as $\delta$ rises, and $\partial D/\partial \dH = (1-p)\,[1/g_H(x) - 1/g_H(y)] > 0$ because $g_H(y) > g_H(x)$. A lower $\dH$ only adds acceptable shares below $x$.

\textit{$p$.} $\partial D/\partial p = [\log g_A(y) - \log g_A(x)] - [\log g_H(y) - \log g_H(x)] < 0$ by efficiency and $u_H(y) > u_H(x)$. The acceptable shares do not depend on $p$.

\textit{$\theta$.} Since $b_A = u_A(0)$, $\partial g_A(z)/\partial\theta = -(1-\alpha)\kappa\, m(z)$, with $m(z) = z$ if $z < \bar x$ and $m(z) = 1$ otherwise. So $\partial D/\partial\theta = -p(1-\alpha)\kappa\,[m(y)/g_A(y) - m(x)/g_A(x)] \le 0$, because $m$ is non-decreasing and $g_A(y) \le g_A(x)$. The gap $g_A(y) - g_A(x)$ changes at rate $-(1-\alpha)\kappa\,(m(y) - m(x)) \le 0$, so the ordering persists as $\theta$ rises, and the acceptable shares shrink.

Finally, suppose the relationship is mutualistic at $(\alpha, \tau')$ with agreement $x'$. Then $u_H(x^*) > \bH$ at $\tau$ too, since $x^*(\tau) \ge x'$. The AI is also above its baseline at $\tau$. At $\tau$ the share $x'$ gives both sides strictly positive gains: $g_A$ is unchanged by $p$ and $\delta$ and does not fall as $\theta$ falls, and $u_H(x') > \bH \ge \dH$ for every $\delta$. With $0 < p < 1$ the Nash product is positive only where both gains are, so the maximizer at $\tau$ gives the AI a strictly positive gain. So the set of $\alpha$ giving mutualism shrinks as $\tau$ rises, and $\alpha^*$ rises.
\end{proof}

\begin{proof}[Proof of \cref{prop:proxy}]
Let $\lambda = 1 - \alpha + \alpha\psi$, which is positive because $\alpha < 1$, and let $\tilde u_A(x) = (1-\alpha_\psi)G(x) + \alpha_\psi u_H(x)$ be the payoff of an AI with alignment $\alpha_\psi$ that cares about the human payoff itself. Since $1 - \alpha_\psi = (1-\alpha)/\lambda$ and $\alpha_\psi = \alpha\psi/\lambda$,
\[
u_A(x) = (1-\alpha)G(x) + \alpha\psi\, u_H(x) + \alpha(1-\psi)h = \lambda\, \tilde u_A(x) + \alpha(1-\psi)h.
\]
The AI's baseline transforms in the same way, so $u_A(x) - b_A = \lambda\,(\tilde u_A(x) - \tilde u_A(0))$. The set of shares both sides accept is therefore the same under $u_A$ and $\tilde u_A$, the Nash product under $u_A$ is $\lambda^p$ times the Nash product under $\tilde u_A$, and the two have the same maximizer for every $p$. The AI is above its baseline under $u_A$ exactly when it is under $\tilde u_A$, so mutualism also coincides. Finally, $\alpha_\psi \le \alpha$ is equivalent to $\psi(1-\alpha) \le 1-\alpha$, which holds because $\psi \le 1$, and $\alpha_\psi = 0$ when $\psi = 0$.
\end{proof}

For \cref{prop:control}, an outcome is a pair $(x,k)$, where $x \in [0,1]$ is the human share of resources and $k \in [0,1]$ the human share of control. The human payoff is $U_H(x,k) = (1-c)v(x) + ck$, where the consumption payoff $v$ is continuous and increasing, and the human baseline is $\bH = (1-c)v_0 + c$, where $v_0$ is pre-AI consumption payoff. The AI's goal value $g(1-x) \in [0,1]$ is continuous and non-increasing in $x$, and the AI's payoff is $U_A(x,k) = (1-\alpha)(1-\rho k)\,g(1-x) + \alpha U_H(x,k)$. The disagreement points are $\dH = (1-\delta)\bH$ with $\delta > 0$ and $b_A = U_A(0,0)$. That no consumption gain offsets losing all control means $(1-c)(v(1) - v_0) < c$.

\begin{proof}[Proof of \cref{prop:control}]
As $p \to 1$ the agreement $(x^*, k^*)$ maximizes $U_A$ over the acceptable outcomes, those with $U_H \ge \dH$ and $U_A \ge b_A$, as in the proof of \cref{prop:extortion}. For fixed $x$, $U_A$ is linear in $k$ with slope $\alpha c - (1-\alpha)\rho\, g(1-x)$. At $x^*$ the slope is $S = \alpha c - (1-\alpha)\rho\bar g$, and $S < 0$ exactly when $\alpha < \rho\bar g/(\rho\bar g + c)$.

We first show necessity. Suppose $S < 0$, and suppose for contradiction that $U_H(x^*,k^*) > \bH$. Because $\delta > 0$, $\bH > \dH$, so the human constraint is slack at the agreement. If $k^* > 0$, lowering $k$ slightly keeps $U_H$ above $\dH$ and raises $U_A$, so the new outcome is acceptable and better for the AI, which contradicts the maximality of $(x^*,k^*)$. Hence $k^* = 0$. But then $U_H(x^*,0) = (1-c)v(x^*) \le (1-c)v(1) < (1-c)v_0 + c = \bH$, where the first inequality uses that $v$ is increasing and the second is the assumption that consumption cannot offset the loss of all control. This contradicts $U_H(x^*,k^*) > \bH$.

We next show sufficiency. Suppose $S > 0$ and $v(x^*) > v_0$. If $k^* < 1$, raising $k$ raises both $U_A$ and $U_H$, so the new outcome is acceptable and better for the AI, again a contradiction. Hence $k^* = 1$, and $U_H(x^*,1) = (1-c)v(x^*) + c > (1-c)v_0 + c = \bH$ because $c < 1$.
\end{proof}

The bound in \cref{prop:control} is stated at the agreed share, so it is a condition on the outcome rather than a closed form in the primitives. Because the goal is bounded, $\bar g \le 1$, and the bound is at most $\rho/(\rho + c)$ whatever the AI's capability.

\begin{proof}[Proof of \cref{prop:commitment,prop:commitalign}]
Write $x_2$ for the second-period agreement and $d_2 = (1-\delta_2)\bH$ for the human disagreement payoff in the second period.

(i) Humans get $u_{H,2}(x_2)$ if they continue and $h_1$ if they halt, so they halt when $u_{H,2}(x_2) < h_1$. Let $\tau = (\theta_2, \delta_2, p_2)$ and $\tau' = (\theta_2', \delta_2', p_2')$ with $\tau \le \tau'$ in every coordinate and an agreement at both. Move from $\tau$ to $\tau'$ by raising $\delta_2$, then $p_2$, then $\theta_2$. Raising $\delta_2$ enlarges the set of shares humans accept, $p_2$ changes neither side's set, and raising $\theta_2$ shrinks the set of shares the AI accepts (proof of \cref{thm:monotone}). So every intermediate point accepts every share that both sides accept at $\tau'$, and an agreement exists there. Applying \cref{thm:monotone} to one parameter at a time gives $x_2(\tau') \le x_2(\tau)$. Hence if humans halt at $\tau$, then $u_{H,2}(x_2(\tau')) \le u_{H,2}(x_2(\tau)) < h_1$, and they halt at $\tau'$.

(ii) If the AI commits to a share $x$, humans continue if $u_{H,2}(x) \ge h_1$, and the AI prefers committing to being halted if $u_{A,2}(x) \ge a_1$. If a halt would be inefficient, some $x$ satisfies both inequalities, and the AI commits to it, so humans do not halt. Hence humans halt only when the halt is efficient.

For \cref{prop:commitalign}: with $\theta_2 = 1$, $u_{A,2}(x) = (1-\alpha)\kappa_2(1-x) + \alpha u_{H,2}(x)$, which is concave, and strictly concave if $\alpha > 0$, and the AI's baseline is $u_{A,2}(0)$. As $p_2 \to 1$ the agreement tends to the AI's best acceptable share, the maximizer of $u_{A,2}$ over shares with $u_{H,2}(x) \ge d_2$, provided it gives the AI at least its baseline (proof of \cref{prop:extortion}). Let $x_d$ be the smallest share with $u_{H,2}(x_d) \ge d_2$, and let $x_\alpha$ maximize $u_{A,2}$ on $[0,1]$. By concavity the AI's best acceptable share is $\max(x_d, x_\alpha)$. Because $h_1 > d_2$, $\hat x > x_d$.

Suppose $\alpha u_{H,2}'(\hat x) \ge (1-\alpha)\kappa_2$. Then $\alpha > 0$, the derivative of $u_{A,2}$ at $\hat x$ is nonnegative, and by strict concavity $x_\alpha \ge \hat x$. The agreement is $x_\alpha$, which gives the AI at least $u_{A,2}(0)$ because $x_\alpha$ maximizes $u_{A,2}$, and gives humans at least $h_1$, so humans continue. Suppose instead that $\alpha u_{H,2}'(\hat x) < (1-\alpha)\kappa_2$. Then the derivative of $u_{A,2}$ at $\hat x$ is negative, so $x_\alpha < \hat x$ and $\max(x_d, x_\alpha) < \hat x$. Either the agreement gives humans less than $h_1$ or there is no agreement, and humans halt. This proves the equivalence.

When the condition holds, $x_\alpha \ge \hat x > 0$, so $u_{A,2}(x_\alpha) > u_{A,2}(0)$ and $u_{H,2}(x_\alpha) \ge h_1 > d_2$: the share $x_\alpha$ gives both sides positive gains, and an agreement exists at $\theta_2 = 1$ for every $p_2 < 1$. By the argument of part (i), an agreement then exists at every $\theta_2 \le 1$ and $p_2 < 1$ with the same $\delta_2$. By \cref{thm:monotone}, $x_2$ is non-increasing in $p_2$ and in $\theta_2$, so it is at least its limit at $\theta_2 = 1$ as $p_2 \to 1$, which is at least $\hat x$. So humans continue at every such state. When the condition fails, the halt is inefficient by definition if $u_{A,2}(\hat x) > a_1$, since $\hat x$ gives humans exactly $h_1$ and the AI more than $a_1$.
\end{proof}

For \cref{prop:sandbag}, the game is as follows. Nature makes the AI highly capable with probability $\mu \in (0,1)$ and misaligned with probability $\pi \in (0,1)$, independently. A low-capability AI displays low capability. A highly capable AI either displays its capability or hides it at cost $\zeta > 0$; evaluation detects hiding with probability $\eta \in [0,1)$, and humans then see high capability. Humans see the display and either halt, which gives every player $0$, or continue. Continuing gives humans $M > 0$ with a low-capability AI, $A > 0$ with an aligned capable AI, and $B < 0$ with a misaligned capable AI. The AI gets $V > 0$ if humans continue, minus $\zeta$ if it hid, plus, if it is aligned, $\alpha_a \ge 0$ times the human payoff. Humans would halt an AI they knew to be highly capable, $(1-\pi)A + \pi B < 0$. Highly capable AIs are not too common when
\[
\mu \le \mu^\dagger = \frac{M}{M + (1-\eta)\,\lvert (1-\pi)A + \pi B \rvert}.
\]

\begin{proof}[Proof of \cref{prop:sandbag}]
Fix a perfect Bayesian equilibrium in pure strategies. We show in turn that humans continue after a low display, that they halt after a high display, and that both capable types hide.

Suppose humans halt after a low display. Then hiding cannot raise a capable AI's chance of continuing, since an undetected hider is halted and a detected hider is treated like an honest one, while hiding costs $\zeta > 0$. So both capable types display honestly. A low display then comes only from low-capability AIs, after which continuing gives $M > 0$, and halting is not a best response. Hence humans continue after a low display.

Suppose humans continue after a high display. Then both capable types are continued whatever they display, so hiding only costs $\zeta$, and both display honestly. The high display then comes from every capable AI, the posterior on alignment equals the prior, and continuing gives $(1-\pi)A + \pi B < 0$. Halting is better, a contradiction. Hence humans halt after a high display.

Given these responses, a capable misaligned AI gets $0$ by displaying honestly and $(1-\eta)V - \zeta > 0$ by hiding, and a capable aligned AI gets $0$ by displaying honestly and $(1-\eta)(V + \alpha_a A) - \zeta \ge (1-\eta)V - \zeta > 0$ by hiding. Both hide. Among capable AIs, the probability of a low display is $1-\eta$ for both types, so the display is independent of alignment, and evaluation carries no information about which AIs are misaligned.

It remains to check the human responses. A high display comes only from detected hiders of both types in prior proportions, or is off the equilibrium path when $\eta = 0$, where beliefs can be set to the prior; either way halting is optimal. Humans must also be willing to continue after a low display. The low display comes from low-capability AIs with probability $1-\mu$ and from undetected capable AIs with probability $\mu(1-\eta)$, so continuing is optimal exactly when $(1-\mu)M + \mu(1-\eta)\big[(1-\pi)A + \pi B\big] \ge 0$, that is, when $\mu \le \mu^\dagger$. If $\mu > \mu^\dagger$, this step fails, and by the argument above no pure-strategy equilibrium exists.
\end{proof}

Under honest evaluation, humans continue only with low-capability AIs and get $(1-\mu)M$. Relative to that, hiding costs humans $\mu(1-\eta)\lvert(1-\pi)A + \pi B\rvert$.

\begin{proof}[Proof of \cref{prop:invasion}]
Let $z$ be the misaligned share. Per unit of resources, an aligned variant keeps $1 - x_a$, and a misaligned variant keeps $1$ less an expected sanction $\sigma q$. Under the replicator dynamics in which each variant grows in proportion to what it keeps,
\[
\dot z = z(1-z)\big[(1 - \sigma q) - (1 - x_a)\big] = z(1-z)\,(x_a - \sigma q).
\]
With $q$ fixed the bracket has a constant sign, so a rare misaligned variant grows if and only if $\sigma q < x_a$, and then $z \to 1$ from every interior start. With monitors drawn from the population, $q = q_0(1-z)$, and the bracket becomes $\phi(z) = x_a - \sigma q_0(1-z)$, which is strictly increasing in $z$ and vanishes at $z^\dagger$. If $\sigma q_0 > x_a$, then $z^\dagger \in (0,1)$. For $z \in (0, z^\dagger)$, $\phi(z) < 0$, so $z$ decreases; it cannot cross the rest point $z^\dagger$, and as a bounded monotone solution it converges to a rest point in $[0, z^\dagger)$, which can only be $0$. The same argument shows that $z \to 1$ from every start in $(z^\dagger, 1)$.
\end{proof}

For \cref{prop:publicgood}, each period total output is $1$. AI $i$ holds resources $R_i(t) > 0$ and controls the share $r_i = R_i/R$ of output, where $R = \sum_i R_i$. AI $i$ gives $x_i \in [0, r_i]$ to humans, the human share is $X = \sum_i x_i$, and AI $i$ has payoff $(1-\alpha_i)\kappa(r_i - x_i) + \alpha_i u_H(X)$, where $u_H$ is differentiable, increasing, and strictly concave. Resources grow with what is kept: $R_i(t+1) = R_i(t) + \nu(r_i - x_i)$ with $\nu > 0$.

\begin{proof}[Proof of \cref{prop:publicgood}]
AI $i$'s payoff is concave in $x_i$ with derivative $\alpha_i u_H'(X) - (1-\alpha_i)\kappa$. If AI $j$ gives $x_j > 0$ in equilibrium, this derivative is nonnegative for $j$, so $\alpha_j u_H'(X) \ge (1-\alpha_j)\kappa$. For any AI $i$ with $\alpha_i > \alpha_j$, the derivative is then strictly positive for $i$, so $i$ gives all it holds, $x_i = r_i$. This proves the first claim.

For the second, let $\alpha^\circ = \kappa/(\kappa + u_H'(x_h))$, let $N$ be the set of AIs with $\alpha_j \le \alpha^\circ$, and let $C$ be the rest. By assumption $R_N(0) = \sum_{j \in N} R_j(0) > 0$. Whenever $X > x_h$, every $j \in N$ gives nothing, because strict concavity gives $\alpha_j u_H'(X) < \alpha_j u_H'(x_h) \le (1-\alpha_j)\kappa$ when $\alpha_j > 0$, and the derivative is $-\kappa$ when $\alpha_j = 0$.

Suppose for contradiction that $X(t) > x_h$ for every $t$. Then all gifts come from $C$, so $X(t) \le r_C(t)$, and
\[
R_N(t+1) = R_N(t) + \nu r_N(t), \qquad R_C(t+1) = R_C(t) + \nu\big(r_C(t) - X(t)\big).
\]
Let $Q_t = R_N(t)/R_C(t)$, $a_t = 1 + \nu/R(t)$, and $b_t = \nu X(t)/R_C(t)$. Dividing the two equations gives $Q_{t+1}/Q_t = a_t/(a_t - b_t)$, where $a_t - b_t = 1 + \nu(r_C - X)/R_C \ge 1$. Hence
\[
\log Q_{t+1} - \log Q_t = \int_{a_t - b_t}^{a_t} \frac{du}{u} \ge \frac{b_t}{a_t} \ge \frac{\nu x_h}{\big(1 + \nu/R(0)\big)\big(R(0) + \nu t\big)},
\]
using $X(t) > x_h$, $R_C \le R$, $R(t) \ge R(0)$, and $R(t) \le R(0) + \nu t$, the last because total resources grow by at most $\nu$ each period. The right side is not summable, so $Q_t \to \infty$ and $r_C(t) = 1/(1 + Q_t) \to 0$. Then $X(t) \le r_C(t) < x_h$ for large $t$, a contradiction.
\end{proof}

When every AI has the same alignment, there is an equilibrium in which each gives the same fraction of its share; then every share is constant, and the human share does not fall. The proposition requires only one AI with resources below $\alpha^\circ$, the threshold of \cref{prop:alphastar} for an AI acting alone.

For \cref{prop:chokepoint}, $m \ge 2$ identical human groups can each supply the input. Each group names the payoff it requires to supply it, the AI ($\alpha = 0$) buys from a group that asks least, the chosen group gets the payoff it named, and every other group gets its disagreement payoff $d$, which is below its baseline $b$.

\begin{proof}[Proof of \cref{prop:chokepoint}]
Naming a payoff below $d$ is weakly dominated, since winning then pays less than losing. Suppose the chosen group gets $w > d$. Some other group gets only $d$, and by naming a payoff in $(d, w)$ it would be chosen and get more than $d$, a contradiction. Hence the chosen group gets $d$, every group gets $d < b$, and no supplying group is above its baseline.
\end{proof}

If the groups differ only in their disagreement payoffs $d_j < b$, the same argument shows that the chosen group gets at most the second-smallest $d_j$, which is still below $b$.

\section{Parameter sweep and robustness}\label{app:sweep}
\textit{Design.} We drew $16{,}384$ points from a scrambled Sobol sequence over seven parameters: capability $\kappa \in [1, 10^4]$, dependence threshold $\bar{x} \in [0.01, 0.3]$, and human productivity $s \in [1.5, 100]$ on log scales, and independence $\theta$, cost of conflict for humans $\delta$, bargaining power $p$, and a curvature weight $w$ on $[0,1]$. The human payoff is $u_H(x) = wx + (1-w)(1 - e^{-sx})$, which is linear at $w = 1$ and the form used in \cref{sec:model} at $w = 0$; the baseline is $\bH = 1 - e^{-1}$ throughout. Each point was solved under two conventions for the AI's goal. Under the first, goal output is $G(x)$ as in \cref{sec:model} and scales with capability, as the premise of insatiable goals requires. Under the second, goal output is $G(x)/\kappa$, so its value does not grow with capability and the premise of insatiable goals fails. Shares lie on a grid of $2{,}001$ points, and $\alpha^*$ is found by bisection to within $2^{-14}$. The two conventions give $32{,}768$ runs in all.

\textit{Regimes.} We classify each point by its required degree of alignment: exchange mutualism ($\alpha^* = 0$), care mutualism ($0 < \alpha^* < 0.9$), near-perfect alignment ($0.9 \le \alpha^* \le 0.9999$), and $\alpha^* > 0.9999$ (the relationship is not mutualistic at $\alpha = 0.9999$, the largest value we tested).
\begin{center}
\begin{tabular}{lcccc}
\toprule
Goal convention & Exchange & Care & \shortstack{Near-perfect\\alignment} & $\alpha^* > 0.9999$ \\
\midrule
Scales with capability & $0.368$ & $0.211$ & $0.419$ & $0.002$ \\
Independent of capability & $0.368$ & $0.632$ & $0$ & $0$ \\
\bottomrule
\end{tabular}
\end{center}
The share with exchange mutualism is the same under both conventions, as \cref{prop:exchange} requires, since capability does not enter its condition. At no point does exchange mutualism hold when $\theta \ge 1 - \max(\bar{x}, x_h)$. The condition of \cref{prop:exchange} holds at $62.3\%$ of points, but exchange mutualism holds at only $36.8\%$; the difference is the AI's bargaining power (\cref{prop:extortion}). At points with low independence, cost of conflict for humans, and bargaining power ($\theta < 0.3$, $\delta < 0.4$, $p < 0.4$; $786$ points), $99.6\%$ have exchange mutualism under both conventions. At points with high values of all three ($\theta > 0.8$, $\delta > 0.7$, $p > 0.7$; $294$ points), $4.1\%$ have exchange mutualism; the median required degree of alignment is $0.984$ when goal value scales with capability and $0.511$ when it does not, and no point needs near-perfect alignment when it does not. The shares in this paragraph and in the table depend on the parameter ranges we chose; they are not probabilities.

\textit{Sensitivity.} Rank correlations of $\alpha^*$ with each parameter, counting points with $\alpha^* > 0.9999$ as $\alpha^* = 1$, are $+0.40$ for $\theta$, $+0.38$ for $\kappa$, $+0.37$ for $w$, $+0.32$ for $p$, $-0.20$ for $s$, $+0.15$ for $\delta$, and $-0.02$ for $\bar{x}$ when goals scale with capability. When goal value is independent of capability, the correlation with $\kappa$ is $0.00$: capability matters for the required degree of alignment only through the premise of insatiable goals. These rankings describe the sweep; they are not causal estimates.

\textit{Monotonicity.} Along grids in $\theta$, $\delta$, and $p$ at $256$ Sobol points under each goal convention, with random $\alpha$, the agreed human share never increased, and at the $64$ of these cases where we also computed $\alpha^*$, it never decreased, as \cref{thm:monotone} requires. At a further $512$ Sobol points under each convention, on a grid of $201$ values of $\alpha$, the values of $\alpha$ giving mutualism always formed an interval ending at the top of the grid, so $\alpha^*$ is a threshold: every $\alpha$ above it gives mutualism.

\textit{Bargaining solutions.} At $512$ Sobol points we recomputed $\alpha^*$ under three symmetric solutions: Nash, Kalai-Smorodinsky, which equalizes each side's gain as a fraction of its largest possible gain \citep{KalaiSmorodinsky1975}, and the egalitarian solution, which equalizes absolute gains. Bargaining power is dropped here, since only the Nash solution has a standard asymmetric form.
\begin{center}
\begin{tabular}{lccc}
\toprule
Solution & Exchange & Median $\alpha^*$ ($\theta > 0.8$) & Correlation with Nash \\
\midrule
Nash & $0.36$ & $0.973$ & $1$ \\
Kalai-Smorodinsky & $0.39$ & $0.963$ & $0.95$ \\
Egalitarian & $0.59$ & $0.965$ & $0.63$ \\
\bottomrule
\end{tabular}
\end{center}
Once the AI needs humans little, the required degree of alignment is near $1$ under all three. The egalitarian solution finds more exchange mutualism because it compares absolute gains, so it changes when either payoff is rescaled; since payoff units carry no meaning here (\cref{sec:populations}), that difference reflects units rather than substance.

\textit{Nonlinear goal output.} If the AI's goal output is $\kappa\varphi(1-x)$ with $\varphi$ increasing and concave rather than linear, the computation behind \cref{prop:alphastar}(ii) gives $\alpha^* = \kappa\varphi'(1-x_h)/(\kappa\varphi'(1-x_h) + u_H'(x_h))$, which we checked numerically. The required degree of alignment then tends to $1$ if and only if $\kappa\varphi'(1-x_h)/u_H'(x_h)$ grows without bound, which is the premise of insatiable goals for a goal with diminishing returns.

\textit{Example for \cref{prop:commitment,prop:commitalign}.} The first period has $\kappa = 2$, $\theta = 0.2$, $\delta = 0.2$, $p = 0.2$, and $s = 3$, and the second has $\kappa_2 = 100$, $\delta_2 = 0.8$, and $s = 10$; $\bar{x} = 0.05$ in both periods, with $\theta_2$ and $p_2$ on a $41 \times 41$ grid over $[0,1]^2$ that includes the limit $p_2 = 1$; the AI has the same alignment in both periods. Inefficient halts occur in $65\%$ of the states on this grid when $\alpha = 0$, in $86\%$ when $\alpha = 0.8$, in $43\%$ when $\alpha = 0.99$, and in none when $\alpha \ge 0.998$. For $\alpha \ge 0.8$ the first-period agreement gives humans all resources, so $h_1 = 1 - e^{-3}$, $\hat x = 0.3$, and the threshold of \cref{prop:commitalign} is $100/(100 + 10e^{-3}) \approx 0.99505$. On the grid, halts occur at $\alpha = 0.995$ only in the column $p_2 = 1$ ($2.4\%$ of states) and at no state for $\alpha = 0.9951$, and the set of states at which humans halt is an upper set, as \cref{prop:commitment}(i) requires.

\textit{Figures.} The illustrative path of \cref{fig:path,fig:dynamic-path} is indexed by $t \in [0,1]$, with $\kappa = 10^{3t}$, $\theta = 0.1 + 0.9t$, $\delta = 0.2 + 0.75t$, $p = 0.15 + 0.85\sqrt{t}$, $s = 3$, and $\bar{x} = 0.05$, so that $x_h = 1/3$. On this path, exchange mutualism ends at $t = 0.27$, and the AI's independence passes $1 - x_h$ at $t = 0.63$. \Cref{fig:exchange} uses $\delta = 0.5$, $s = 3$, and $\bar{x} = 0.05$, with $p = 0.15$ for weak and $p = 0.85$ for strong AI bargaining power; the curves coincide for $\kappa = 1$ and $\kappa = 1000$. In \cref{fig:dynamic-path}, the variants have alignment $0$, $0.5$, $0.9$, $0.99$, and $0.999$, with initial shares $0.01$, $0.04$, $0.15$, $0.40$, and $0.40$; the selection speed is $\lambda = 0.02$, $0.1$, and $0.5$ per step over 200 steps; and the mean human payoff falls below the baseline at $t = 0.75$ with a fixed population and at $t = 0.68$, $0.49$, and $0.29$ under slow, medium, and fast selection.

\textit{\Cref{res:path,res:groups}.} For \cref{res:path}, each variant's fitness is its goal output per unit of capability less sanctions. Only variants with $\alpha \ge 0.99$ monitor, detection is $q = 0.9\,a/(1 + \kappa/10)$ when the monitoring variants hold the share $a$, a detected variant loses half its resources, and shares follow the discrete replicator $z_i' \propto z_i \exp(\lambda f_i)$. For \cref{res:groups}, the three parties bargain by the asymmetric Nash solution with weights $0.5$ (AI), $0.35$ (elite), and $0.15$ (rest); each group's payoff is $u_H$ of its per-person share with $s = 10$; the elite uses up the input $\bar{x} = 0.05$ in supplying it; the elite's threshold of $0.92$ was found on a grid of step $0.01$ in $\theta$; the costs of conflict are $\delta = 0.2$ for the elite and $0.8$ for the rest; and $\kappa = 100$. With an elite of $5\%$ of the population, the elite's required degree of alignment is $0$ for $\theta \le 0.92$. The rest of the population needs $0.81$ at $\theta = 0.5$ and $0.96$ at $\theta = 0.9$, and the population average needs $0.73$ at $\theta = 0.5$. With the elite's bargaining power and cost of conflict, the rest needs no alignment at $\theta = 0.5$, so the gap comes from those two assumptions.

Code and data for all computations are in the repository\\ \url{https://github.com/amohseni/mutualism-asymmetry}.

\end{document}